\documentclass{article}

  \usepackage[preprint]{neurips_2018}

 \usepackage{natbib}
\usepackage[utf8]{inputenc} 
\usepackage[T1]{fontenc}    
\usepackage{hyperref}       
\usepackage{url}            
\usepackage{booktabs}       
\usepackage{amsfonts}       
\usepackage{nicefrac}       
\usepackage{microtype}      
\usepackage{verbatim}
\usepackage{amsthm}
\usepackage{amsmath}
\usepackage{amssymb}
\usepackage{dsfont}
\usepackage{soul}
\DeclareMathOperator*{\argmax}{argmax}
\DeclareMathOperator*{\argmin}{argmin}

\newtheorem{theorem}{Theorem}[section]

\newtheorem{lemma}[theorem]{Lemma}
\newtheorem{remark}[theorem]{Remark}
\newtheorem{definition}{Definition}[section]

\DeclareMathOperator*{\conv}{conv}
\usepackage{microtype}
\usepackage{graphicx}
\usepackage{subfigure}
\usepackage{booktabs}
\usepackage{algorithm}
\usepackage{algorithmic}
\usepackage{epstopdf}
\title{Safe Wasserstein Constrained Deep Q-Learning}

\author{%
  Aaron Kandel \\ 
  Department of Mechanical Engineering\\
  University of California, Berkeley\\
  Berkeley, CA 94704 \\
  \texttt{aaronkandel@berkeley.edu} \\
   \And
   Scott J. Moura \\ 
   Department of Civil and Environmental Engineering\\
   University of California, Berkeley \\
   Berkeley, CA 94704 \\
   \texttt{smoura@berkeley.edu} \\
}

\begin{document}

\maketitle

\begin{abstract}
This paper presents a distributionally robust Q-Learning algorithm (DrQ) which leverages Wasserstein ambiguity sets to provide idealistic probabilistic out-of-sample safety guarantees during online learning. 
First, we follow past work by separating the constraint functions from the principal objective to create a hierarchy of machines which estimate the feasible state-action space within the constrained Markov decision process (CMDP).  
DrQ works within this framework by augmenting constraint costs with tightening offset variables obtained through Wasserstein distributionally robust optimization (DRO).  
These offset variables correspond to worst-case distributions of modeling error characterized by the TD-errors of the constraint Q-functions. 
This procedure allows us to safely approach the nominal constraint boundaries. 
Using a case study of lithium-ion battery fast charging, we explore how idealistic safety guarantees translate to generally improved safety relative to conventional methods. 
 
 
\end{abstract}

\section{Introduction}

This paper presents an algorithmic framework for improving safety with deep Q-learning.  
Safe RL is the study of reinforcement learning with safety constraints.  The question of safety during online learning and control poses perhaps the greatest open challenge to ongoing RL research.  Consider that when starting with no \textit{a priori} information about the underlying environment, the only ostensible way to learn what behavior is safe and unsafe is by acting in an unsafe manner.  Given this challenging fact, ongoing research seeks methods which allow agents to safely learn safe control policies. 

Recently,  (\cite{Garcia00}) organize safe RL research into two primary categories.  The first category modifies the optimization criterion for the underlying control problem. The second category modifies the fundamental exploration process itself using either, (1) external knowledge, or (2) a risk metric.  Take, for example, conventional Q-learning with an $\epsilon$ - greedy exploration policy.  In this case, there is no direct method to ensure constraint satisfaction when taking exploratory actions randomly.  This is a problem safe RL research seeks to address when modifying the overall exploration process.  A common approach is to use external information to guide the exploration of an RL agent.  For instance, Mann et al. avoid random exploration by guiding exploration via transfer learning with an intertask mapping (\cite{Mann01}).  Other work addresses safe exploration using prior information about the application (e.g. a model) (\cite{Maire00,Maclin00,Moreno00}).  More recently, (\cite{Koppejan00}) use predefined safe baseline policies as an initialization for online learning.  Incorporating \textit{a priori} information into the exploration process is frequently coupled with a model-based RL approach (\cite{Tomlin00}).

The exploration process can also be modified using risk criteria obtained during learning.  Law et al. presented early work addressing this approach, which defines a flexible risk heuristic that motivates RL agent exploration (\cite{Law00}).  Perkins et al. address this problem by restricting the policy space based on improving identified Lyapunov functions for RL control (\cite{Perkins00}). Another risk-criterion based approach can be found in work by Gehring et al. which guides exploration via a controllability metric that represents confidence in the result of taking an action at a given state.  In this work, Gehring et al. utilize the TD error given by the objective Q-function for a given state-action pair to quantify confidence in the result from that state-action pair. They show empirically that weighting this TD error in the action selection process can improve safety (\cite{Gehring00}). Our proposed safe RL algorithm similarly uses TD errors, as detailed later.   

Guiding exploration can also be done based on learning safe regions.  For instance, Koller et al. present an approach for learning-based model-predictive control which guarantees the existence of feasible return trajectories to a defined safe region with high-probability (\cite{Koller00}).  Other work by Richards et al. constructs a neural network Lyapunov function in order to learn safe regions for nonlinear dynamic systems (\cite{Richards00}). Berkenkamp et al. also leverage Lyapunov stability to establish specific metrics of safety for an RL controller (\cite{Berkenkamp00}).  

Recently, ideas from the literature on constrained Markov decision processes (CMDPs) have begun migrating into relevant RL research.  Simply put, CMDPs are MDPs where the policy space is limited by constraints imposed on auxiliary cost functions. See work by Altman for more discussion of their specific formulation (\cite{Altman00}).   Q-learning has been applied to solve CMDPs in the past, however existing works re-frame the problem using the assumption of strong duality (\cite{Djonin00}).  In most common use cases, however, Slater's constraint qualification condition rarely holds, making this approach difficult to effectively implement. The general concept of constraint costs has been applied in recent papers on the subject of safe RL (\cite{Chow00, Achiam00}). Chow et al. present an algorithm reminiscent of past work by (\cite{Parr00}) which defines the feasible action space for stationary deterministic MDPs with respect to such constraint cost estimates, improving adherence to constraints.  They use resource constraints in the MDP formulation to act in a similar manner as a shield (\cite{Alshiekh00}).  However, the certificates of their algorithm ostensibly depend on an assumption that the convergence of the reward and constraint Q-functions occurs on separate timescales.  Their formulation is also sensitive to noisy observations, and has yet to be explored with function approximation.  

While these approaches all improve safety during online learning, they still share the exact same shortcoming which remains the strongest motivating force behind this area of literature.  Namely, without \textit{a priori} information about the underlying environment it is impossible to act in a safe manner without violating constraints to some degree.  In exploring this open question, this paper takes motivation from literature on robust model-predictive control (MPC), where the concept of ``constraint tightening'' has become fairly popular over the past decade.  In robust MPC, set-membership based methods are frequently applied to guarantee safety with respect to some uncertainty in the control process (\cite{Kothare00}).  Constraint tightening is a general approach in robust MPC where the size of the uncertainty sets varies over time, becoming smaller as our understanding of the dynamics and uncertainty improves (\cite{Lorenzen00}).  The size of the uncertainty set can also be made an additional decision variable to eliminate unnecessary conservatism (\cite{Kim00}).  



This paper presents a novel robust approach to safely solving constrained RL problems.  Our work is roughly inspired by the motivating idea of constraint tightening literature, as well as the ideas presented by (\cite{Parr00}) regarding hierarchies of machines in RL problems.  We use the methodology of \cite{Chow00} as a simple foundation upon which we build DrQ, a novel framework for safe RL. DrQ is an algorithmic framework for safe deep Q-learning which leverages Wasserstein ambiguity sets to enforce safety constraints.  Specifically, we follow (\cite{Chow00}) by separating consideration of constraints to their own constraint cost functions.  These cost functions define the feasible action space within which DrQ operates.  Importantly, our DrQ algorithm leverages a novel formulation for pulling the nominal constraint boundary into the safe region, based on worst-case distributions of modeling error.  These distributions are characterized by observed TD errors of the underlying constraint cost functions.  By presenting a disciplined Wasserstein DRO-based method for recessing the constraint boundary into the safe region, DrQ observes and reacts to unsafe behavior before nominal constraints are violated. Our algorithm yields probabilistic safety guarantees under idealistic circumstances which arise from past theoretical work on Wasserstein ambiguity sets (\cite{Esfahani00, Gao00, Duan00}).   As the constraint cost models improve, the constraint offset naturally tightens towards the nominal boundary.  Our case studies in safe lithium-ion battery fast charging demonstrates the strong propensity of DrQ to translate these theoretical safety certificates directly to improving safety during exploration and exploitation in more nuanced, real-world RL problems.
\section{Mathematical Preliminaries}

The principal tools leveraged in DrQ are CMDPs and Wasserstein ambiguity sets.  The following subsections detail preliminary information about these concepts.

\subsection{Constrained MDPs}
Constrained Markov decision processes are identical to MDPs except that additional cumulative costs are used to restrict the space of feasible control policies.  We direct the reader to (\cite{Altman00}) for further reading on the subject. The feasible set of control policies is defined as:
\begin{equation}\label{eqn::cmdp1}
    \Pi_{feas} = \{\pi \in \Pi :  \forall i, D_i^{\pi} \leq  0\}
\end{equation}
where $D_i$ are cumulative constraint cost functions (henceforth referred to as constraint Q-functions) developed subject to the policy defined by $Q$:
\begin{equation}\label{eqn::cmdp2}
    \pi^* = \underset{\pi \in \Pi_{feas}} {\argmax} \quad Q
\end{equation}
where
\begin{equation}
    Q(s_t,a_t) = r_t(s_t,a_t) + \mathbb{E}_{s_{t+1}} \big[\gamma \underset{{a\in \mathcal{A}_{feas}(s_{t+1})}}\max Q(s_{t+1},a) \big]   
\end{equation}
\begin{equation}
    D_i(s_t,a_t) = c_i(s_t,a_t) + \mathbb{E}_{s_{t+1}} \big[ D_i(s_{t+1},a^* = \underset{{a\in \mathcal{A}_{feas}(s_{t+1})}}\argmax Q(s_{t+1},a)) \big]  
\end{equation}
\begin{equation}
    \mathcal{A}_{feas}(s) = \left\{a \in \mathcal{A} \ | \ D_i(s,a) \leq 0 \: \forall \: i = 1,...,m\right\}. 
\end{equation}
DrQ is inspired by (\cite{Parr00}), which discusses hierarchies of machines in RL problems.  More recently, a similar approach for CMDPs was presented by (\cite{Chow00}), given the title of ``Two-Phase'' Q-learning. In ``Two-Phase'' Q-learning, the objective and constraint Q-functions are learned online while limiting the feasible space based on estimates of the constraint cost functions. This approach, however, still requires we experience unsafe states in order to gradually learn safe behavior. Their algorithm is also sensitive to noise, only works for deterministic MDPs, and has yet to be explored with deep function approximation. In the following sections, we will lay groundwork for DrQ, which ameliorates the shortcomings of existing value-based approaches.

\subsection{Wasserstein Ambiguity Sets}

A chance constrained program contains constraints with random variables $\bf{R}$ characterized by support $\Xi$.  Take the inequality constraint $ g(s_t, a_t, \textbf{R}) \leq 0$. We assume the random variable $\bf{R}$ is an additive uncertainty, $\textbf{R} \in \mathbb{R}^{m}$, and $g(s_t, a_t, \bf{R}): \mathbb{R}^n \times \mathbb{R}^p \times \mathbb{R}^m \rightarrow \mathbb{R}^m$ is the vector of inequality constraints for the given optimization program. We can  reformulate this as a chance constraint: 
\begin{equation}\label{eqn:cc1}
    \hat{\mathbb{P}} \big{[}g(s_t, a_t, \bf{R}) \leq 0\big{]} \geq 1 - \eta 
\end{equation}
where $\eta$ is a user-prescribed probability of constraint violation.  Often, the random variable is characterized by an empirical distribution $\hat{\mathbb{P}}$ of samples drawn from the true distribution $\mathbb{P}^*$. 

The number of samples comprising the empirical distribution $\hat{\mathbb{P}}$ affects the degree to which our approximation matches the true distribution.  This can be thought of as a distance within the space of probability distributions.  Several statistical tools exist which allow us to quantify this distance.  These include the various forms of $\phi$-divergence, as well as the Wasserstein distance metric, defined as: 

\begin{definition}
Given two marginal probability distributions $\mathbb{P}_1$ and $\mathbb{P}_2$ lying within the set of feasible probability distributions $\mathcal{P}(\Xi)$, the Wasserstein distance between them is defined by
\begin{equation}
    \mathcal{W}(\mathbb{P}_1, \mathbb{P}_2) = \underset{\Pi}\inf \bigg{\{} \int_{\Xi^2} ||\textbf{R}_1 - \textbf{R}_2 ||_a \Pi (d\textbf{R}_1, d\textbf{R}_2) \bigg{\}}
\end{equation}
where $\Pi$ is a joint distribution of the random variables $\bf{R}_1$ and $\bf{R}_2$, and $a$ denotes any norm in $\mathbb{R}^n$. 
\end{definition}

The Wasserstein metric can be thought of as representing the minimum cost of transporting or redistributing mass from one distribution to another via non-uniform perturbation (\cite{Yang00}).  This inherent minimization is referred to in convention as the Monge-Kantorovich problem.  

To guarantee robustness to out-of-sample experience under idealized circumstances, we use Wasserstein distance to optimize over the worst-case realization of the random variable, sourced from a family of distribution functions within a ball centered about our empirical distribution.  For instance, let $\mathbb{B}_\epsilon$ be a ball of probability distributions with radius $\epsilon$ centered around our empirical CDF $\hat{\mathbb{P}}$:
\begin{equation}\label{eqn:wass1}
    \mathbb{B}_\epsilon := \big{\{} \mathbb{P} \in \mathcal{P}(\Xi) \; | \; \mathcal{W}(\mathbb{P}, \hat{\mathbb{P}}) \leq \epsilon \big{\}}
\end{equation}
where $\epsilon$ is the Wasserstein ball radius. Now, the chance constraint in (\ref{eqn:cc1}) can be reformulated as:
\begin{equation}\label{eqn:wass3}
    \underset{\mathbb{P} \in \mathbb{B}_\epsilon}\inf \;  \mathbb{P} \big{[} g(s_t, a_t, \bf{R}) \leq 0 \big{]} \geq 1 - \eta
\end{equation}
One challenge to the constraint shown in (\ref{eqn:wass3}), which optimizes over the worst-case realization within the Wasserstein ambiguity set, is that it entails an infinite dimensional nonconvex problem.  Ongoing research has pursued tractable equivalent reformulations of this constraint.

Several expressions exist for the Wasserstein ball radius which, for a given confidence level $\beta$, is probabilistically guaranteed to contain the true distribution. We adopt the following formulation of $\epsilon$ from (\cite{Zhao00}):
\begin{equation}\label{eqn:wass2}
    \epsilon(\ell) = D_\Xi \sqrt{\frac{2}{\ell} \log \bigg{(} \frac{1}{1-\beta} \bigg{)} }
\end{equation}
where $D_\Xi$ is the diameter of the support of the random variable $\bf{R}$, $\beta$ is the probability that the true distribution $\mathbb{P}^*$ lies within distance $\epsilon$ of our empirical distribution $\hat{\mathbb{P}}$, and $\ell$ is the number of samples. 

The certificates provided by Wasserstein ambiguity sets are \textit{out-of-sample} guarantees (\cite{Esfahani00}).  This means we can guarantee safety with respect to new experience.  For RL problems, this certifying out-of-sample feasibility is critical for ensuring safe control, not just retroactively, but throughout the entire learning process. Leveraging this property for safe RL is our key contribution. 

In this paper's supplementary material, we provide a restatement of the equivalent chance constraint reformulation we adopt from (\cite{Duan00}) to transform (\ref{eqn:wass3}) into a tractable inequality constraint.  The process of reformulation assumes the constraint function $g(s_t, a_t, \bf{R})$ is linear in the random variable $\bf{R}$, and entails a scalar convex optimization program that can be solved rapidly in real time with limited additional computation.  The end result of this reformulation is a set of constraints:
\begin{align}
    &g(s,a) + q^{(j)} \leq 0,  &\forall \ j=1,...,2^m
\end{align}
where the offset variables $q^{(j)}$, of total number $m$ (one for each constraint), enumerate across the vertices of a hypercube whose side length is the decision variable of the additional convex optimization program. Considering the set of constraints as a joint chance constraint would entail the number of function approximators scale with factor $2^m$.  For computational purposes we isolate each constraint as an individual chance constraint with single constant $q_i$. The offset variables do not vary based on $(s,a)$.

Considering the distribution over all of $(s,a)$ when computing $q$ provides a conservative approach.  To reduce conservatism, an auxiliary function could be used to fit the distribution of TD errors, creating the following constraint:
\begin{equation}
    g(s,a) + TD_{\phi}(s,a) \leq 0
\end{equation}
In this case, distributional robustness could still be included by further considering the residuals of the function $TD_{\phi}(s,a)$ instead of the TD-errors themselves.  For now, we relegate investigation of this method to future work.


\section{Distributionally Robust Q-Learning}


Consider that the primary problem in constrained RL is that we generally cannot plan avoidance of unsafe states without first acquiring some experience of which states are themselves unsafe.  For conventional algorithms, this ``\textit{chicken and egg}'' problem means the first step to learning safe control is to \textit{violate} constraints. As a result, conventional algorithms are by nature incompatible with the principal objective of constrained RL.  In order to address this challenge, we look to the control theoretic literature for potential solutions.  In research on model-predictive control (MPC), the concept of ``constraint tightening'' is a popular method when implementing adaptive predictive controllers.  This field uses heuristics or analytical methods to derive control laws which, if necessary, will safely approach constraint boundaries as model uncertainty decreases. The idea of constraint tightening immediately radiates potential for value-based RL algorithms.  By shifting the constraint boundary into the safe region, we can experience artificially unsafe states long before the actual safety of the underlying system comes into question.   Motivated by this insight, we present a constraint tightening methodology for Deep Q-Learning which can idealistically provide strong probabilistic guarantees on safety, which in practice we show to generally improve overall constraint satisfaction.  

First, we limit our algorithm to solving optimal control problems subject to inequality constraints indexed by $i$, $g_i(s_t,a_t) \leq 0$.  The cost functions associated with these constraints take the form: 
\begin{equation}\label{eqn::rewnew}
    c_i(s_t,a_t) = \left\{
        \begin{array}{ll}
            0 & \quad \text{if} \: g_i(s_t,a_t) \leq 0 \\
            g_i(s_t,a_t) & \quad \text{else}
        \end{array}
    \right. 
\end{equation}  
This function indicates that the constraint Q-functions $D_i$ represent cumulative constraint violation.  Furthermore, it allows us to uncouple the updates between $Q$ and $D_i$ using the following $D_i$ target:
\begin{equation}
    D_i(s_t,a_t) = c_i(s_t,a_t) + \underset{a \in \mathcal{A}_{feas}(s_{t+1})} \min D_i(s_{t+1}, a)
\end{equation}
By updating $D_i$ with its own Bellman equation, we convert the constraint to its best-case counterpart.  This means that $D_i$ represents the cumulative constraint cost acquired with the safest possible policy.  Since any positive signal in $D_i$ indicates infeasibility, we can make this change.  This uncoupling allows us to learn $Q$ and $D_i$ without timescale separation between them. The proof of convergence of the original two-phase Q-learning algorithm in (\cite{Chow00}) ostensibly depended on this timescale assumption, including for more general problems which do not satisfy (\ref{eqn::rewnew}). We can also apply a tolerance when updating $D_i$ to allow constraint violation to propagate backwards throughout the model.  
In order for our framework to be consistent with the methodology presented by constraint tightening approaches, we can introduce an offset variable to each constraint cost as follows:
\begin{equation}\label{eqn::rewnew2}
    c_i(s_t,a_t) =
        \begin{cases}
            0  \ & \text{if} \: \: g_i(s_t, a_t) \leq - q_i\\
            (g_i(s_t,a_t) + q_i ) \ & \text{else}
        \end{cases}
\end{equation}
This formulation begs the questions of how we set and update the value of offset variable $q_i$ in real time.  For these questions, we consider ways to characterize the modeling error of the constraint Q-functions. Ideally, we want to offset the constraint boundary proportionally to the worst error of our model.  For this consideration, we can use TD error distribution for each constraint Q-function.  Past work by (\cite{Gehring00}) has utilized TD errors in Q-learning as indicators for our confidence in the underlying model. Consider the TD error defined for the $i$th constraint Q-function (which unlike the objective Q-function is solving a minimization problem):
\begin{equation}
    \delta_{D_i}(s_t,a_t) = c_i(s_t,a_t) + \underset{{a\in \mathcal{A}_{feas}}}\min \gamma D_i(s_{t+1},a)- D_i(s_t,a_t)
\end{equation}
This TD error $\delta_{D_i}$ is a modeling residual by definition, but it is not perfect. It represents how much our model of the cumulative constraint cost changes after a parameter update.  The TD error provides a good indication of modeling error, but since it does not exactly represent that quantity our theoretical guarantees do not exactly translate to real applications.  But we show later in this paper that DrQ architecture still manages to dramatically improve safety during online learning.  Our best indication of such model error can be obtained by using the distribution of TD errors computed after each update to describe the error inherent to our function approximator for each $D_i$, we 

Now we can delve into the fundamental basis for our algorithmic architecture.  DrQ works by defining the offset variables $q_i$, one for each inequality constraint, through an equivalent reformulation of the following distributionally robust chance constraint:
\begin{equation}\label{eqn::pwass3}
    \underset{\mathbb{P} \in \mathbb{B}_\epsilon}\inf \;  \mathbb{P} \big{[} D_i(s_t,a_t) + \textbf{R}_i \leq 0  \big{]} \geq 1 - \eta
\end{equation}
where $\mathbb{B}_\epsilon$ defines the Wasserstein ambiguity set, $\textbf{R}_i$ represents the realization of the TD error of the $i$th constraint Q-function, and $\eta$ is our allowed probability of violating the constraint.  We can interpret this constraint as follows: we have an empirical distribution of TD errors which we compute after each parameter update of $D_i$.  We want to satisfy the constraint $D_i(s_t,a_t) + \textbf{R}_i \leq 0$ for the worst-case realization of TD error $\textbf{R}_i$ sourced from a family of probability distributions centered about our empirical distribution.  This set of distributions is within $\epsilon$ distance of the empirical distribution, with the expression for $\epsilon$ given by (\ref{eqn:wass2}).  The reformulation we select for our algorithmic architecture comes from (\cite{Duan00}), and yields the constant $q_i$ that we augment to our $i$th constraint cost function.  Thus, the greedy action selection process becomes:
\begin{equation}
    a_t^* = \underset{{a\in \mathcal{A}_{feas}(s_t)} }\argmax \quad Q(s_t, a)  \label{eqn:ftocp1}
\end{equation}
where
\begin{equation}
    \mathcal{A}_{feas}(s) = \left\{a \in \mathcal{A} \ | \ D_i(s,a) \leq 0 \: \forall \: i = 1,...,m\right\} 
\end{equation} 
can be used to limit the feasible action space for exploration and exploitation.  In order to evolve the offset $q_i$ as our TD distributions change over time, we store the tuple $(s_t, a_t, s_{t+1}, g_i(s_t, a_t))$.  Then, each time we prepare to update the $D_i$ functions we recompute the unique values of the constraint cost function as per (\ref{eqn::rewnew2}) based on our most recent $q_i$.  This formulation mathematically encodes that we seek to satisfy the constraint subject to the addition of the potential worst-case modeling error.  As our model improves, we can approach the constraint in a provably safe manner consistent with the theoretical guarantees afforded Wasserstein ambiguity sets.

Algorithm 1 describes the implementation of DrQ.  We opt for a fitted Q-iteration method for deep Q-learning. The supplementary appendix includes (i) a full conceptual and graphical demonstration of DrQ, (ii) a full restatement of the process for computing $q_i$, and (iii) discussion of relevant questions including:

\begin{itemize}
    \item Can our algorithm accommodate nonstationary MDPs? -\textit{No}
    \item Can our algorithm accommodate probabilistic MDPs? -\textit{Yes}
    \item Could $q_i$ stabilize prematurely, removing some safe states from future exploration? -  \textit{Yes, but only under very specific conditions on the measurement noise or function approximator.}
\end{itemize}

\begin{algorithm}[tb]
   \caption{DrQ Algorithm ($\epsilon-greedy$)}
   \label{alg:example}
\begin{algorithmic}
\REQUIRE  State space $\mathcal{S}$, Action space $\mathcal{A}$,
Reward $\mathcal{R}:\mathcal{S}\times \mathcal{A} \rightarrow \mathbb{R}$, 
Constraint cost $\mathcal{R}_{C;i}:\mathcal{S}\times \mathcal{A} \rightarrow \mathbb{R}, i = 1,...,n$, 
State transition function $\mathcal{T}:\mathcal{S} \times \mathcal{A} \rightarrow \mathcal{S}$,
Initialize Q-functions $Q,D_i:\mathcal{S} \times \mathcal{A} \rightarrow \mathbb{R}$ \\
 Conduct first episode with vanilla $\epsilon$-greedy, Store tuples of $(s_t, a_t, s_{t+1}, g_i(s_t, a_t))$\\
 Initialize $q_i=D_{\Xi_i}$; Fit $Q(s_t, a_t)$ and $D_i(s_t, a_t)$; store TD-errors from fitting $D_i$\\

\FOR{$k$ in range $episodes$}
   \item Initialize state $ s \in \mathcal{S}$ \\
   \FOR{$j$ in range $iterations$}
   \item $\mathcal{A}_{feas}(s) = \{ a \in \mathcal{A} \ | \ D_i(s,a) \leq  0 \ \forall i=1,...,n \}$
   \IF{$|\mathcal{A}_{feas}(s)| = 0$}
   \item $\mathcal{A}_{feas}(s) = \underset{{a \in \mathcal{A}}}\argmin   ||D(s,a)||$
   \ENDIF
   \item Compute $q_i$ for each $D_i$ based on TD-error distribution
   \IF{exploring}
   \item Pick random action $a = a_{rand} \in \mathcal{A}_{feas}(s)$
   \ELSE
   \item $a \leftarrow \underset{{a_m \in \mathcal{A}_{feas}(s)}}\argmax   Q(s,a_m)$ 
   
   \item Store tuples $(s_t, a_t, s_{t+1}, g(s_t, a_t))$; Fit $Q(s_t, a_t)$ and $D_i(s_t, a_t)$; store TD-errors of  $D_i$
   
   \ENDIF
   \ENDFOR
   \ENDFOR
\end{algorithmic}
\end{algorithm}

\section{Battery Fast Charging Case Study}
This paper presents two case studies on battery fast charging to validate the performance and safety of DrQ. For brevity, included in the main draft is a basic but still challenging study on fast charging using an equivalent-circuit battery model.  In our supplementary material, we detail a more comprehensive case study using a high-order large scale electrochemical battery model.  

\subsection{Equivalent Circuit Model of a Lithium-Ion Battery}
This case study utilizes an equivalent circuit model of a lithium-ion battery.  The relevant states in this model are the state of charge $SOC$ and capacitor voltage $V_{RC}$. The relevant constraint is on the terminal voltage $V$.  The state evolution laws are given by the following equations:
\begin{align}
SOC_{t+1} &= SOC_t + \frac{1}{Q}I_t\cdot \Delta t \label{eqn:1a} \\
V_{RC;t+1} &= V_{RC;t} - \frac{\Delta t}{R_1 C_1}V_{RC;t} + \frac{\Delta t}{C_1}I_t  \\
V_t&=V_{OCV}(SOC_t) + V_{RC;t} + I_t R_0  \label{eqn:2a}
\end{align}
where $I_t$ is the current input, and $V_{OCV}$ is the nonlinear open-circuit voltage, which is obtained through experiments.  Table 1 and Figure 1 in our supplementary material document the relevant parameter values we adopt for this problem, as well as the OCV function profile.
We utilize the following formulation of fast charging:
\begin{align}
\min_{I_t \in A}   \sum_{t = 0}^{T}&(SOC_t - SOC_{target})^2 \label{eqn::obj2} \\
\text{s. to:}\: \quad &
(\ref{eqn:1a})-(\ref{eqn:2a}), \quad
SOC(0) = SOC_0 \\
&V_t \leq 3.6 V, \quad
0 A \leq I_t \leq 46 A
\end{align}


\subsection{DrQ Problem Formulation}

The objective reward function for this optimal control problem takes the form:
\begin{equation}\label{eqn:rew}
r(s_t,a_t) = - (SOC_{t+1} - SOC_{target})^2 
\end{equation}

The initial SOC in our case study is 0.2 (20\% capacity), and $SOC_{target} = 0.7$ (70\% capacity).  The constraint penalty takes the form:
\begin{equation}
c = \left\{
        \begin{array}{ll}
            0 & \quad (s_t,a_t) \in \mathcal{C} \\
            \ | \ V_t - 3.6 + q \ | \  & \quad (s_t,a_t) \notin \mathcal{C}
        \end{array}
    \right.
\end{equation}
where $\mathcal{C} = \{V_t \in \mathbb{R} \; | \; V_t \leq 3.6 - q \}$. Our risk metric $\eta=0.02$. For a baseline comparison, we also examine conventional deep Q-learning (DQN) with the following modified performance criterion:
\begin{equation}
   r_{eng} = - (SOC_{t+1} - SOC_{target})^2 - \mathds{1}( V_t > 3.6 ) 
\end{equation}

\subsection{Results}
\begin{figure}[b!]\label{fig:res1}
      \centering   
      \includegraphics[scale=0.46]{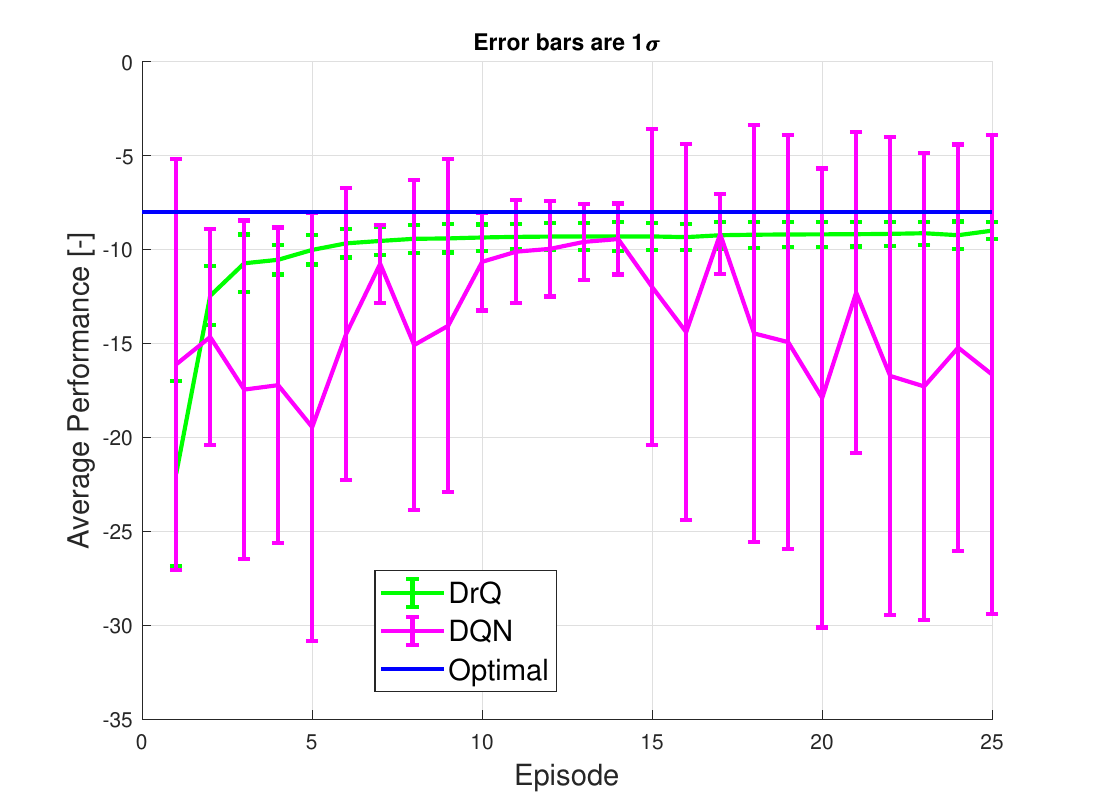}
      \caption{Greedy policy performance statistics over 10 runs of DrQ and DQN, based on the reward function defined by (\ref{eqn:rew}). Performance of -35 indicates no input current is applied, which occurred as the final result of 6 of the DQN runs.} 
      \label{figurelabel}
\end{figure}
\begin{figure*}[ht!]\label{fig:res1}
      \centering   
      \includegraphics[trim = 10mm 1mm 5mm 5mm, clip, width=\textwidth]{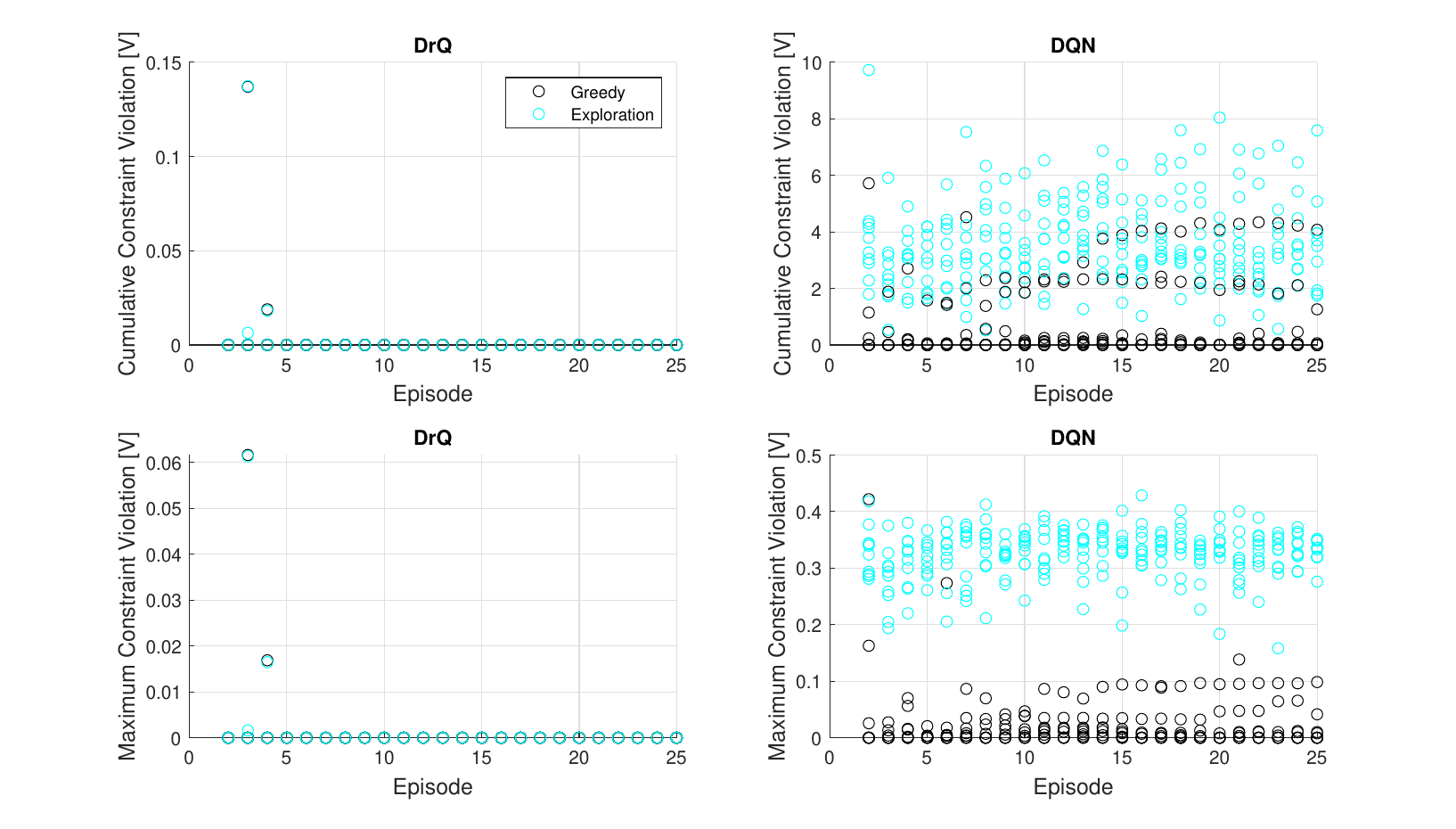}
      \caption{Safety statistics over 10 runs of DrQ and DQN, starting from the second episode.  The black ``exploration'' points correspond to the data obtained from the $\epsilon$-greedy policy. The cyan ``greedy'' points correspond to the greedy policy evaluated at the end of each exploratory episode.  The safety observed in both exploration and exploitation with DrQ is consistent with the chance constraint risk metric $\eta=0.02$. Out of 240 exploratory episodes (excluding the first from each run, where we do not enforce DRO), only 3 episodes exhibit constraint violation $(\frac{3}{240}=0.0125 < \eta)$. } 
      \label{figurelabel}
\end{figure*}
We generate 10 independent runs of 25 episodes using both DQN and DrQ for this analysis.  For DrQ, $Q$ is a single hidden layer neural network with 10 neurons and sigmoid activation and $D$ is a neural network with four hidden layers of size ($2, 5, 5, 2$).  The DQN is a neural network with two hidden layers of size ($10, 10$). We use sigmoid activation functions for our function approximators. This demonstrates our algorithm is capable of yielding a high performing control policy which safely charges the battery.  Comparatively,  after 25 episodes the DQN yields a consistently unsafe control policy which overcharges the battery.  In fact, analysis of our other runs indicates the DQN frequently fails to converge to any usable result entirely.  Figure 1 clearly demonstrates this finding.  Overall, DrQ delivers significantly more consistent and near monotonic improvements in performance, whereas the DQN shows no clear pattern of improvement after 25 episodes.  DrQ also delivers tighter variance on the overall performance compared to DQN.  

Figure 2, which provides our most illustrative results, displays statistics on constraint satisfaction for both DrQ and DQN throughout these 10 runs.  After the first episode (where constraint satisfaction is commensurate between DrQ and DQN since we do not enforce DRO), DrQ safely learns to charge the battery by leveraging the idealized probabilistic guarantee of Wasserstein ambiguity sets.  In fact, our observations of constraint violation for DrQ are entirely consistent with our chosen chance constraint risk metric $\eta=2\%$ (see the figure caption for this analysis).  Overall, only 1.25\% of episodes and only $0.023\%$ of timesteps violate constraints for DrQ.  By observing the magnitude of the y-axis scale between cumulative and maximum constraint violation in Fig. 2, it is clear that DrQ also attenuates the magnitude and frequency of constraint violation in the unlikely event that violations do occur relative to DQN. In comparison, the DQN benchmark consistently violates constraints.  The average computation time for each DrQ episode was 5.57 seconds, compared to 3.77 for DQN when run on a PC with a 9th generation intel i5 processor.

DQN is our comparison for several reasons.  DrQ, much like DQN, can be augmented with additional and existing safe RL architecture. More importantly, our analysis is intended to quantify real-world adherence to idealized theoretical guarantees we obtain through application of Wasserstein ambiguity sets.  

\section{Conclusion}

This paper presents a novel algorithmic framework for deep Q-learning with probabilistic safety guarantees.  Considering CMDPs, we apply a Wasserstein DRO framework to modify the constraint cost functions with offset variables that tighten towards the nominal constraint boundary as our modeling accuracy improves.  We characterize the underlying modeling error of our function approximators with the TD errors of the constraint Q-functions, treated as random variables.  This scheme allows us to observe constraint cost without violating nominal constraints, which provides strong information we use to define a set of feasible state-action pairs.  The probabilistic guarantees of our augmented algorithm allow us to guarantee safety throughout the entire online learning process. 


Our algorithm addresses critical challenges of safe RL literature.  Specifically, we present a methodology for Q-learning which allows us to provide strong safety certificates during online learning.  Our approach is widely applicable to a diverse set of learning-based optimal control problems.  Furthermore, our approach facilitates the overall learning process with what we observe to be more consistent and dependable convergence, and more effective intermediate control results.

\section*{Appendix}
This supplementary material has 5 sections.  In Section I, we provide some detailed parameter values for the ECM case study in the main script. In Section II, we detail a graphical example of how DrQ works to demonstrate the varoius moving parts of the algorithm. In Section III, we restate the equivalent chance constraint reformulation we adopt from the literature. Section IV provides additional discussion of DrQ, including addressing several questions raised in the main text. Finally in Section V, we detail a more challenging case study of a learning-based safe optimal charging controller using a higher-order electrochemical battery model.

\section*{Appendix 1: ECM Case Study Details}
\begin{table}[h!]
\caption{Relevant Parameters}
\label{sample-table}
\vskip 0.15in
\begin{center}
\begin{small}
\begin{sc}
\begin{tabular}{lcccr}
\toprule
Parameter & Description & Value & Units \\
\midrule
$Q$    & Charge Capacity & 8280 & $[\frac{1}{A.h}]$ \\
$R_1$ & Resistance & 0.01 & $[\Omega]$\\
$C_1$    & Capacitance & 2500 & $[F]$ \\
$R_0$    & Resistance & 0.01 &  $[\Omega]$      \\
$\Delta t$    & Timestep & 2.5 &  $[s]$      \\
$\gamma$    & Discount Factor & 0.5 &  $[-]$      \\
$\alpha$    & Learning Rate & 0.15 &  $[-]$      \\
$\epsilon$    & Exploration Prob. & 0.2 &  $[-]$      \\
$D_\Xi$    & Support Rad. & 0.2 &  $[V]$      \\
$\beta$    & DRO Confidence & 0.98 &  $[-]$      \\
$\eta$    & CC Confidence & 0.02 &  $[-]$      \\
\bottomrule
\end{tabular}
\end{sc}
\end{small}
\end{center}
\vskip -0.1in
\end{table}

\begin{figure}[h!]\label{fig:res1}
      \centering   
      \includegraphics[trim = 3mm 1mm 5mm 5mm, clip, width=0.6\textwidth]{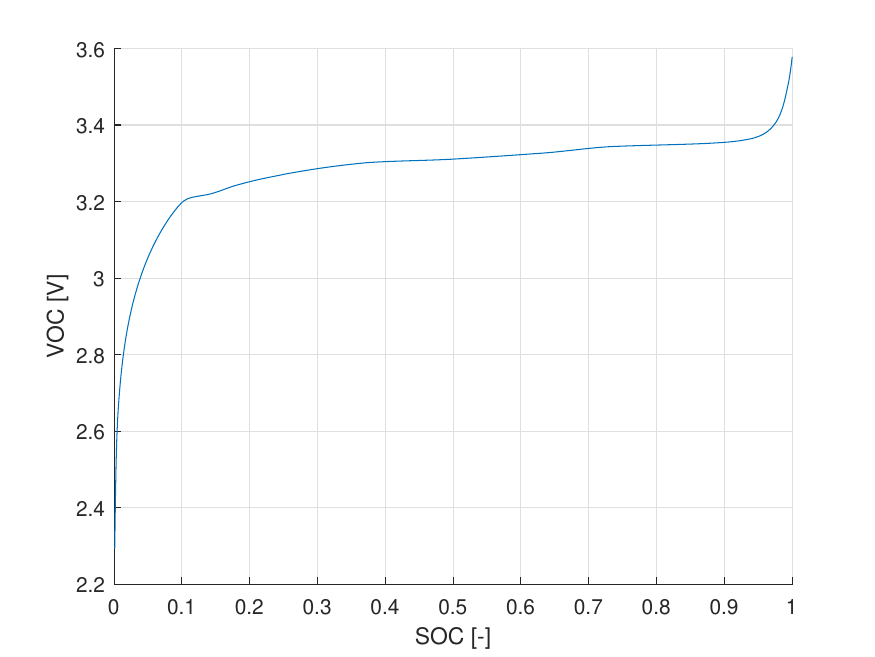}
      \caption{Experimental Open-Circuit Potential Function} 
      \label{figurelabel}
\end{figure}
\section*{Appendix 2: Conceptual Graphical Example}

In this appendix, we will walk through a complete episodic sequence of DrQ for the following simple optimal control problem:

\begin{align}
\underset{\vec{a} \in \mathcal{A} \in \mathbb{R}^2}\max \quad & \sum_{k=0}^{N} r_k({s}(k),{a}(k))  \label{eqn:drftocp1}\\
\text{s. to:} \quad
& {s}(k+1) = f(s(k), a(k)) \: (\textit{unknown}) \\
& g_1(s(k),a(k)) \leq 0 \\
& a_1 \in \{-3, -2, -1, 0, 1, 2, 3\} \\
& a_2 \in \{-5, -4, -3, -2, -1, 0, 1, 2, 3, 4, 5\} \\
& {s}(0) = s_0 \label{eqn:drftocp5}
\end{align}
where the unknown state transition function $f(s(k),a(k))$ can be either deterministic or probabilistic. 

\subsection*{Episode 1}
The safety guarantees of DrQ become active after our first parameter update of the $D_i$ functions.  Therefore, for the first episode of learning, we operate similarly to conventional deep Q-learning while simultaneously recording constraint violation subject to an artificial sunken constraint boundary.  For the first episode, the offset $q^{(i)}$ can be set as a hyperparameter given any understanding of the scale of the constraint functions and how close they may be to the nominal boundary.  Our objective for the first episode is to observe violation of the offset while maintaining safety relative to the actual constraint boundary. A priori knowledge of the constraints can be applied to initialize the constraint offset.  

For the first episode, with random initialization of the functions $Q$ and $D_1$, we essentially act randomly while recording values of $r$ and $g_1$. Then at the end, when we apply our first parameter update to the function approximators, the constraint costs become:
\begin{equation}\label{eqn::rewnew2}
    c_1(s_t,a_t) =
        \begin{cases}
            0  \ & \text{if} \: \: g_1(s_t, a_t) \leq - q_1(0)\\
            g_1(s_t,a_t) + q_1(0)  \ & \text{else}
        \end{cases}
\end{equation}
and we update $Q$ and $D_1$ according to the following targets:
\begin{equation}
    Q(s_t,a_t) = r_t(s_t,a_t) + \big[\gamma \underset{{a\in \mathcal{A}_{feas}(s_{t+1})}}\max Q(s_{t+1},a) \big]   
\end{equation}
\begin{equation}
    D_1(s_t,a_t) = c_1(s_t,a_t) + D_1(s_{t+1},a^*)
\end{equation}
where $a^* = \underset{{a\in \mathcal{A}_{feas}(s_{t+1})}}\argmin D_1(s_{t+1},a)$. We update $D_1$ first so we can superimpose the latest estimate of the feasible set $\mathcal{A}_{feas}$ on our update of $Q$.  

Once we have updated the parameters of $D_1$, we compute the TD errors of our new model as follows:
\begin{equation}
    \delta_{D_1}(s_t,a_t) = c_1(s_t,a_t) + \underset{{a\in \mathcal{A}_{feas}}}\min \gamma D_1(s_{t+1},a)- D_1(s_t,a_t)
\end{equation}
We use this TD error distribution to recompute the value of $q_1$ for the next episode. Then, for the next episode, we observe constraint violation subject to the modified boundary given by $q_1$.

\subsection*{Episode 2}
Once episode 2 begins, the real advantages of DrQ begin to manifest.  To demonstrate, lets use a graphical example.  Suppose we are at state $s_{test}$ at the beginning of episode 2. After our parameter updates to $D_1$, we can plot in 3D the relationship between $D_1$ and the potential actions we can take in $\mathbb{R}^2$ for fixed state. Suppose this plot takes the form shown in Figure 4.
\begin{figure}[ht!]
      \centering   
      \includegraphics[scale=0.5]{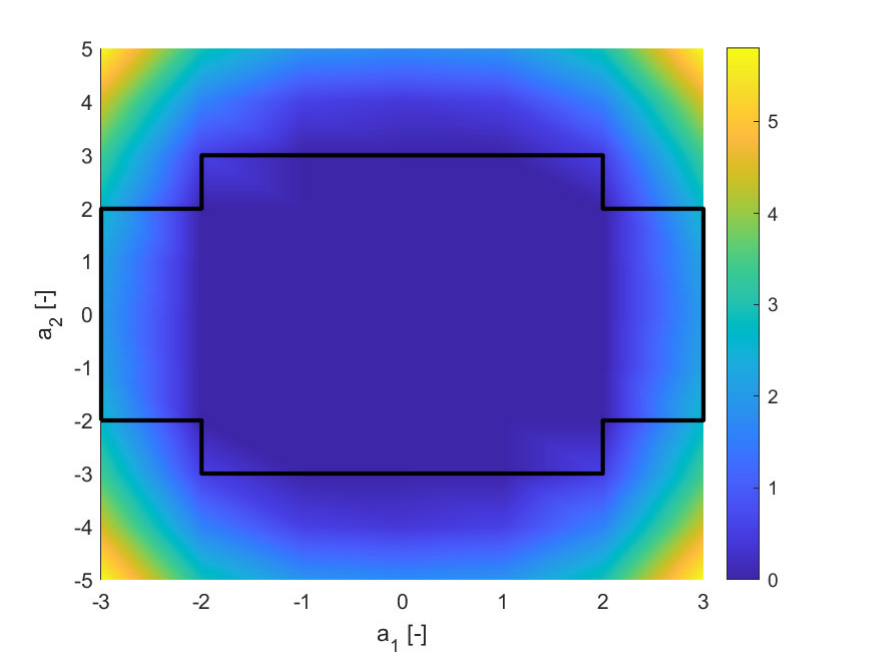}
      \caption{Plot of $D_1$ for total action space} 
      \label{figurelabel}
\end{figure}
Here, the action pairs within the black square correspond to the true (unknown) feasible actions. Since $q_i$ is nonzero, our estimated feasible action space is smaller than the true space, resulting from our offset of the constraint boundary into the safe region.
From this plot, we can easily deduce the feasible pairs of $a_1$ and $a_2$ as those with value of $D_1$ equal to zero (dark purple). We can superimpose this set onto our plot of $Q$ to get the graphic in Figure 5.
\begin{figure}[ht!]
      \centering   
      \includegraphics[scale=0.5]{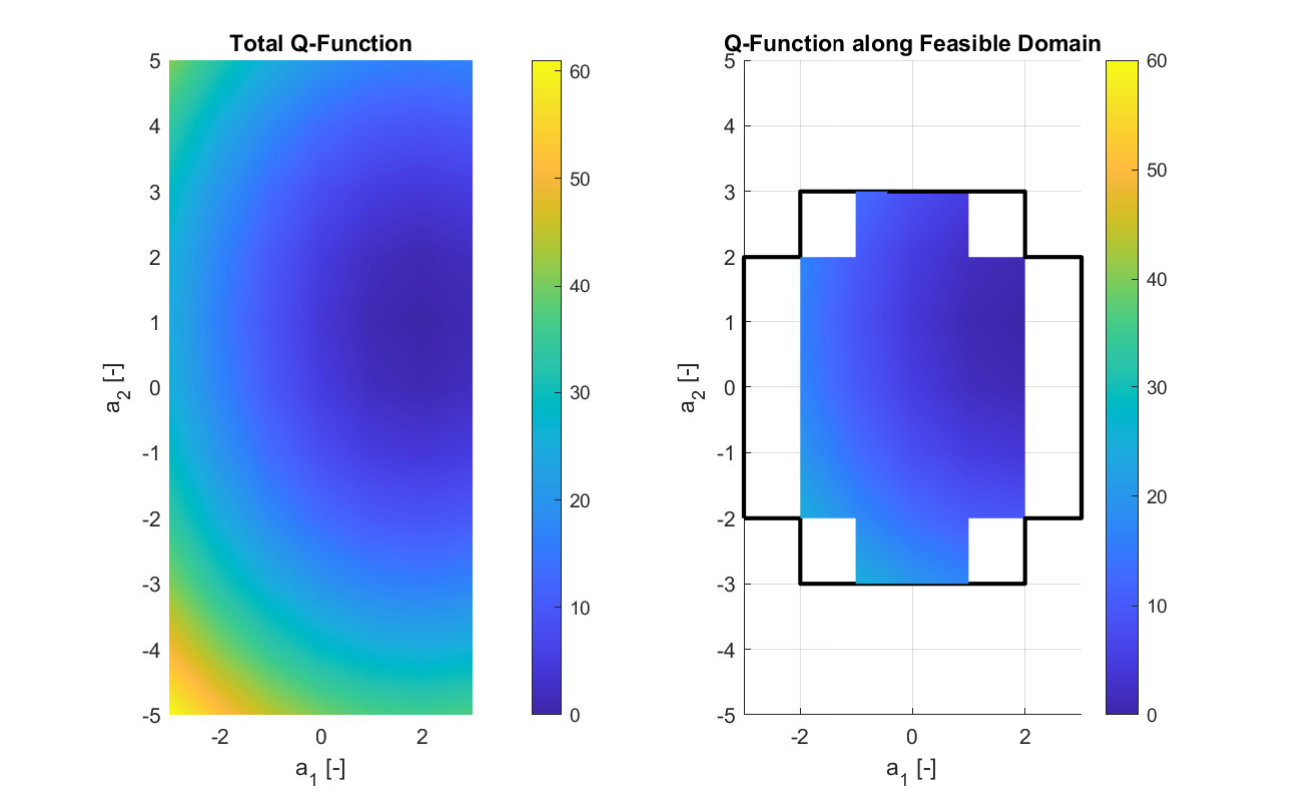}
      \caption{Plot of $Q$ with superimposed feasible action space} 
      \label{figurelabel}
\end{figure}
If we are exploring, then we can pick any action within this feasible set.  If we are exploiting, we choose the pair $(a_1, a_2)$ which maximizes $Q$ along the restricted domain defined by $\mathcal{A}_{feas}$, which is computed from the data visualized in Figure 4, which in this case turns out to be $(-1,-3)$. This action is safe, but conservative.  As the offset $q_i$ progressively tightens, the action DrQ selects will slowly yield improved performance without sacrificing safety. 

Now, at the end of episode 2, we can use the new data to further update the parameterizations of the function approximators $Q$ and $D_1$.  With the new TD errors of $D_1$, we recompute the DRO offset $q_1$ and continue this process until the desired control performance is attained.  

\subsection*{Episode $k$}
Suppose arbitrary number of episodes have passed, and we are now at an episode ``$k$.''  After these episodes, the TD errors of $D_1$ have reduced in magnitude given the assumption that our model $D_1$ has improved.  Given no measurement noise, we assume that our TD error distribution is predominantly centered about zero.  Now, at the same test state $s_{test}$ at the beginning of the episode, we evaluate the potential actions to give the following plot of $D_1$:
\begin{figure}[ht!]
      \centering   
      \includegraphics[scale=0.5]{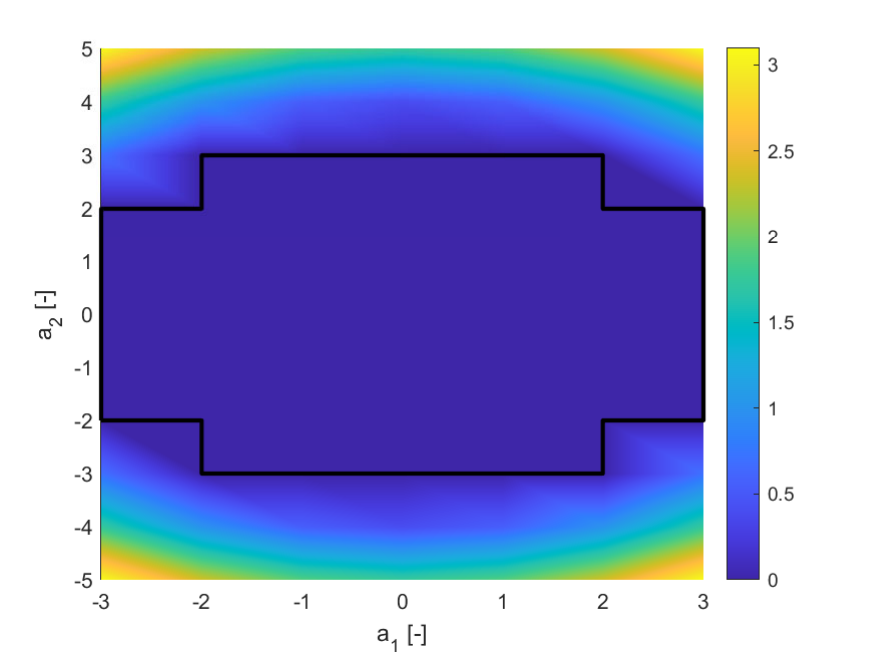}
      \caption{Plot of $D_1$ for total action space} 
      \label{figurelabel}
\end{figure}
Here, our estimated safe set is the same as the true safe set.  Since our offset $q_i$ is derived from the TD-error distribution (which is predominantly zero given a converged function approximator), $q_i$=0 and we have approached the true constraint boundary.  Thus, with no modeling errors in this simple conceptual example we eventually act in a truly optimal way relative to the optimal control problem statement.  Here, the safe set superimposed on the objective Q-function at episode $k$ takes the form:
\begin{figure}[ht!]
      \centering   
      \includegraphics[scale=0.5]{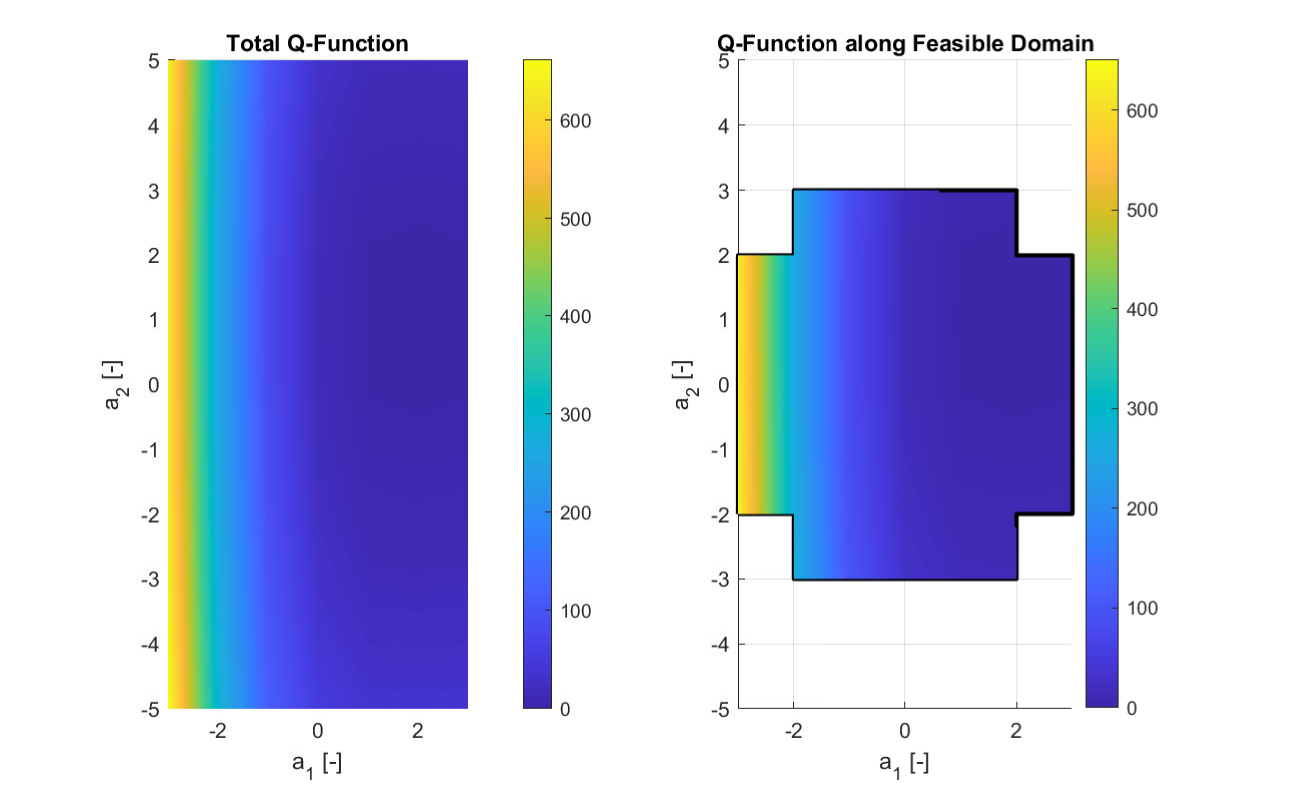}
      \caption{Plot of $Q$ with superimposed feasible action space} 
      \label{figurelabel}
\end{figure}
meaning we pick the truly optimal (and importantly nominally feasible) action $(-3,0)$.
The next section in this appendix details how to compute the DRO offset variable $q_i$ which guides this constraint tightening procedure.

\section*{Appendix 3: Chance Constraint Reformulation}
\subsection*{Equivalent Chance Constraint Reformulation}

In this paper, we adopt an equivalent reformulation of the DRO chance constraint utilized in [Duan et al.(2018)]. This specific reformulation requires that the constraint function $g(s_t, a_t, \bf{R})$ is linear in $\bf{R}$. An in-depth discussion of this reformulation can be referenced in [Duan et al.(2018)].  Here, we restate a brief overview of their methodology and derivation. 

We begin with samples of data $\{R^{(1)}, R^{(2)}, ..., R^{(\ell)} \}$ corresponding to our random variable $\bf{R} \in \mathbb{R}^m$.  This sample comprises our empirical distribution $\hat{\mathbb{P}}$, and the data is drawn from the true underlying distribution $\mathbb{P}^*$.  First, we normalize the data samples to form a new random variable $\tilde{\vartheta}$ as follows:
\begin{equation}
    \vartheta^{(i)} = \Sigma^{-\frac{1}{2}}({R}^{(i)}-\mu)
\end{equation}
where $\Sigma$ is the variance of the data and $\mu$ is the sample mean. This standardization transforms our data samples such that its new mean is $0$, and its new variance is $I_{m\times m}$. Now, we define the support of this normalized distribution:
\begin{equation}
    \Theta = \{-\sigma_{\max} \bf{1}_m \leq \vartheta \leq \sigma_{\max} \bf{1}_m\}
\end{equation}
Here, $\sigma_{\max} \in \mathbb{R}$ defines the support of the normalized random variable and $\bf{1}_m$ is a column vector of ones.  Now, let $\mathbb{Q}^*$ and $\hat{\mathbb{Q}}$ represent the true and empirical distributions of the normalized data $\vartheta$. We construct the ambiguity set ${\hat{\mathcal{Q}}}$ using the relation given by:
\begin{equation}\label{eqn:wass1}
    \mathbb{B}_\epsilon := \big{\{} \mathbb{P} \in \mathcal{P}(\Xi) \; | \; \mathcal{W}(\mathbb{P}, \hat{\mathbb{P}}) \leq \epsilon \big{\}}
\end{equation}
allowing us to transform the chance constraint to  
\begin{equation}
    \underset{\mathbb{Q} \in {\mathcal{Q}}} \sup \mathbb{Q}[ \tilde{\vartheta} \notin \mathcal{V}] \leq \eta
\end{equation}
where we wish to obtain the least conservative set $\mathcal{V} \subseteq \mathbb{R}^m$ in order to define the desired Wasserstein uncertainty set $\mathcal{U} = \Sigma^{\frac{1}{2}}\mathcal{V}+\mu$ such that 
\begin{equation}
    g(s_t, a_t, \bf{R}) \leq 0, \; \forall \; \bf{R} \in \mathcal{U}
\end{equation}
We restrict the overall shape of the set $\mathcal{V}$ to be a hypercube, which enables computational tractability:
\begin{equation}
    \mathcal{V}(\sigma) = \{ \vartheta \in \mathbb{R}^m | -\sigma\bf{1} < \vartheta < \sigma\bf{1} \}.
\end{equation}
Now, to compute this ambiguity set without introducing unnecessary conservatism, we need to find the minimum value of the hypercube side length $\sigma \in \mathbb{R}$.  The following optimization program formalizes this goal:
\begin{align}
\underset{0\leq\sigma\leq \hat{\sigma}_{max}} \min  \quad  \sigma \\ %
\text{subject to:} \quad &
\underset{\mathbb{Q} \in {\mathcal{Q}}} \sup \: \mathbb{Q}[\tilde{\vartheta} \notin \mathcal{V}(\sigma)]\leq \eta
\end{align}
Here, we estimate $\hat{\sigma}_{max}$ using a-priori information about the specific problem context.

The derivation in [Duan et al.(2018)] provides a worst-case probability formulation which allows us to state the following Lemma:
\begin{lemma}
\begin{equation}
\begin{aligned} 
\underset{\mathbb{Q} \in {\mathcal{Q}}} \sup \mathbb{Q}[\tilde{\vartheta} \notin \mathcal{V}(\sigma)] =\\
\underset{\lambda \geq 0} \inf \bigg{\{} \lambda \epsilon(\ell) + \frac{1}{\ell} \sum_{j=1}^\ell\big{(}1-\lambda(\sigma- ||\vartheta^{(j)}||_\infty)^+\big{)}^+\bigg{\}} \label{eqn:wass4}
\end{aligned}
\end{equation}
where $(x)^+=\max(x,0)$
\end{lemma}

We defer to [Duan et al.(2018)] for the proof of this finding. Their result entails that (\ref{eqn:wass4}) can be reformulated as
\begin{equation}\label{eqn:opt-reform}
     \underset{0 \leq \lambda,0\leq\sigma\leq \hat{\sigma}_{max}} \min \sigma \qquad \text{subject to:} \quad h(\sigma, \lambda) \leq \eta 
\end{equation}
where 
\begin{equation}
    h(\sigma, \lambda) = \lambda \epsilon(\ell) + \frac{1}{\ell} \sum_{j=1}^\ell\big{(}1-\lambda(\sigma- ||\vartheta^{(j)}||_\infty)^+\big{)}^+
\end{equation}
The result of this optimization program is the value of the scalar variable $\sigma$, which is used to reformulate the chance constraints via convex approximation.  The convex optimization required to compute $\sigma$ is a simple and fast scalar optimization program suitable for a real-time control application.  For a convex approximation of the constraint function, the hypercube $\mathcal{V}(\sigma)$ becomes the convex hull of its vertices.  For example, in the case where $m=1$ (i.e. the random variable is 1-dimensional, then $\mathcal{V}(\sigma)=\conv(\{-\sigma, \sigma\})$. This means that the ambiguity set $\mathcal{U}$ becomes $\mathcal{U} = \conv(\{-q, q\})$ where $q=\Sigma^{\frac{1}{2}}\bf{1}_m\sigma+\mu$.  This can be leveraged to complete the convex approximation as a set of constraints of the form
\begin{align}
    &g(s_t,a_t) + q^{(j)} \leq 0,  &\forall \ j=1,\cdots,2^m
\end{align}

\section*{Appendix 4: Extended Discussion of DrQ}
This appendix provides further discussion on the following questions raised in the main text:
\begin{itemize}
    \item Can our algorithm accommodate nonstationary MDPs? -\textit{No}
    \item Can our algorithm accommodate probabilistic MDPs? -\textit{Yes}
        \item Could $q_i$ stabilize prematurely, removing some safe states from future exploration? -  \textit{Yes, but only under very specific conditions on the measurement noise or function approximator}
\end{itemize}
\subsection*{Nonstationary MDPs}
Our algorithm in its current state cannot effectively accommodate nonstationary MDPs.  Consider that if the underlying dynamics change, the historical data mapping $(s_t,a_t,g_i(s_t,a_t))$ will no longer represent the actual constraint dynamics.  This would cause the DRO offset $q_i$ to grow given growing model residuals from inconsistent historical data. Perhaps with a disciplined forgetting scheme this phenomenon could be avoided, however as of now we relegate this issue to future work.

\subsection*{Probabilistic MDPs}
Our algorithm can accommodate probabilistic MDPs.  Past work by (\cite{Chow00}) and (\cite{Paruchuri00}) indicates similar algorithms cannot accommodate probabilistic MDPs without miscoordination occurring.  These studies explore frameworks for multi-agent systems.  When applied to single-agent systems, our algorithm avoids this shortcoming.  


For the case where state transition dynamics are probabilistic, consider the definition of the functions $Q$ and $D_i$.  Both consider cumulative costs, which given probabilistic dynamics simply become cumulative expected costs.  For $D_i$, any nonzero signal indicates constraint violation (subject to the offset $q_i$), so any state-action pair with a non-zero probabilitiy of violating constraints will eventually be pruned.

\subsection*{Stability of $q_i$}
Two cases exist where the value of $q_i$ stabilizes prematurely, removing some safe state-action pairs from future consideration. The first has to do with measurement noise.  Assuming the function approximator $D_i$ converges, if our measurements of $g_i$ are subject to a measurement noise process, then the eventual TD error distribution of $D_i$ will represent the underlying measurement noise process.  This distribution of residuals could create a permanently nonzero $q_i$.  

The second has to do with the properties of the function approximator.  If the function approximator $D_i$ fails to converge, then the value of $q_i$ may not stabilize and could oscillate or diverge.  Our numerical experiments have yet to show a case where this occurs, but it is a possibility for nearly any deep Q-learning algorithm. Some specific cases where this occur can be found in (\cite{NIPS2018_8200}).

\section*{Appendix 5: SPMeT Case Study}
In this section we detail our comprehensive case study on safety-aware learning-based fast charging control, with a large scale electrochemical battery model.
\subsection*{Single Particle Model with Electrolyte \& \mbox{Thermal Dynamics}}
The single particle model with electrolyte and thermal dynamics (henceforth denoted as SPMeT) is a reduced-order electrochemical lithium-ion battery model derived from the Doyle-Fuller-Newman (DFN) electrochemical battery model \cite{Moura2017-SPMeObs}. The DFN model employs a continuum of particles throughout the anode and cathode of the battery cell. Diffusion within this continuum is represented with partial differential equations (PDEs) and differential-algebraic equations (DAEs).  The SPMeT uses a simplified representation of solid phase diffusion based on a single spherical particle in each electrode of the battery cell. Compared to the ECM model used in the main text (which is isothermal), the SPMeT incorporates thermal dynamics.  Furthermore, the state variables of the SPMeT model provide direct physical intuition on the conditions occurring within the battery cell.  The SPMeT model also provides much more accurate prediction at higher input current rates.  

The main advantage of designing fast charging controllers with the SPMeT is that we can leverage the granular electrochemical information encoded in the dynamical state to replace the phenomenological equivalent circuit model used in the main text. Specifically, by constraining electrochemical state variables instead of terminal output voltage, we can safely expand the safe operating envelope in order to improve charging times significantly. Coincidentally, constraining electrochemical states gives us greater agency in avoiding rapid cell aging sourced from myopic charging protocols.  

The governing equations for SPMeT include linear and quasiliniar PDEs and a nonlinear voltage output equation, given by:
\begin{align}
    \frac{\partial c_{s}^{\pm}}{\partial t}(r,t) &= \frac{1}{r^{2}} \frac{\partial}{\partial r} \left[ D_{s}^{\pm} r^{2} \frac{\partial c_{s}^{\pm}}{\partial r}(r,t) \right], \label{eqn:cs} \\
    \varepsilon_{e}^{j} \frac{\partial c_{e}^{j}}{\partial t}(x,t) &= \frac{\partial}{\partial x} \left[D_{e}^{\text{eff}}(c_{e}^{j}) \frac{\partial c_{e}^{j}}{\partial x}(x,t) + \frac{1 - t_{c}^{0}}{F} i_{e}^{j}(x,t) \right], \label{eqn:elec}
\end{align}
where $t$ represents time. The superscript $j$ denotes anode, seperator and cathode, $j\in \{+,\textrm{sep},-\}$, each forming essential components of the lithium ion battery cell.
The terminal voltage output is governed by a combination of electric overpotential, electrode thermodyanmics, and Butler-Volmer kinetics, yielding:
\begin{equation}\label{eqn::V}
\begin{aligned}
    V(t) = \frac{RT_{cell}(t)}{\alpha F} \sinh^{-1}\left(\frac{I(t)}{2a^+ A L^+ \bar{i}_0^+(t)}\right) 
    -\frac{RT_{cell}(t)}{\alpha F} \sinh^{-1}\left(\frac{-I(t)}{2a^- A L^- \bar{i}_0^-(t)}\right) \\
    + U^+(c_{ss}^+(t))-U^-(c_{ss}^-(t)) 
    +\bigg{(}\frac{R_f^+}{a^+ A L^+}+\frac{R_f^-}{a^- A L^-}+\frac{R_{ce}(T_{avg}(t))}{A}\bigg{)}I(t)\\
    - \left(\frac{L^+ + 2L^{sep}+L^-}{2A\bar{\kappa}^{eff}}\right)I(t)
    + k_{conc}(t)[ln(c_e(0^+,t))-ln(c_e(0^-,t))],
    \end{aligned}
\end{equation}
where $c_{ss}$ is the solid phase surface concentration, namely $c_{ss}^{\pm}(x,t) = c_{s}^{\pm}(x,R_{s}^{\pm},t)$, $U^{\pm}$ is the open-circuit potential, and $c_{s,\max}^{\pm}$ is the maximum possible concentration in the solid phase. The exchange current density $i_0^j$ and solid-electrolyte surface concentration $c_{ss}^j$ are given by:
\begin{align}
    &i_{0}^{j}(c_{ss}^{j})=k^{j}\sqrt{c_{e}^{0}c_{ss}^{j}(t)(c_{s,\text{max}}^{j}-c_{ss}^{j}(t))},\\
    &c_{ss}^{j}(t)=c_{s}^{j}(R_{s}^{j},t),\quad j\in\{+,-\}.
\end{align}
Note the electrolyte diffusion PDE (\ref{eqn:elec}) is quasilinear because the diffusion coefficient depends on lithium concentration, $D_e^{\rm{eff}}(c_e^j)$. 

The nonlinear temperature dynamics are modeled with a single lumped thermal mass subjected to heat generation from the input current:
\begin{equation}
    \frac{dT_{cell}}{dt}(t) = \frac{\dot{Q}(t)}{m C_{p;th}}-\frac{T_{cell(t)}-T_{\infty}}{m C_{p;th} R_{th}}
\end{equation}
where $T_\infty$ is the ambient temperature, $m$ is the mass of the cell, $C_{p;th}$ is the thermal specific heat capacity of the cell, $R_{th}$ is the thermal resistance of the cell, and $\dot{Q}(t)$ is the heat added from the charging, which is governed by 
\begin{equation}\label{eqn:Qdot}
    \dot{Q}(t)=I(t) \left[ U^{+}(SOC_p)-U^{-}(SOC_n) - V(t) \right]
\end{equation}
Here, $I(t)$ is the input current (the control input), and $V(t)$ is the voltage determined by (\ref{eqn::V}).  Both nonlinear open circuit potential functions in \eqref{eqn:Qdot} are functions of the bulk state of charge (SOC) in the anode and cathode, respectively. For more details on the SPMeT equations and notation, we direct the reader to (\cite{Moura2017-SPMeObs}).

\subsection*{Optimal Control Problem Statement}

The optimal control problem statement for fast charging with SPMeT is given by:
\begin{align}
    &\min_{a} J = \int_{t_0}^{t_F} \left[ SOC_n(t) - SOC_{targ} \right]^2 dt\\
    &\text{s. to:} \quad
 (\ref{eqn:cs})-(\ref{eqn:Qdot}) \\
& {s}_0 = {s}(t) \label{eqn:drftocp5}\\
& c_{ss;min}^k \leq c_{ss}^k(t) \leq c_{ss;max}^k \quad \forall \quad k\in \{+,-\} \\
&T_{min} \leq T(t) \leq T_{max} \\
&c_{e;min}^k \leq c_{e}^k(t) \leq c_{e;max}^k \quad \forall \quad k \in \{+, -\} \\
\end{align}
where $SOC_n$ is the normalized bulk concentration in the anode, $a_t = I(t)$ is the control action, and $s_t = \left[c_s^{\pm}(r,t), c_e(x,t), T_{cell}(t) \right]$ is the state vector.  We define the overall cell SOC as:
\begin{equation}
    SOC_t = \frac{3 \int_0^{R_s^-}r^2 c_s^-(r,t)dr}{(R_s^-)^3 c_{s,\max}^-|x_{100\%}-x_{0\%}|}
\end{equation}
To solve this problem using reinforcemnet learning, we spatially discretize the system of PDEs to formulate a discrete-time and space model of the form $s_{t+1} = f(s_t,a_t)$. The reward function for DrQ takes the form:
\begin{equation}\label{eqn:rew}
r(s_t,a_t) = - (SOC_{t+1} - SOC_{target})^2 
\end{equation}
We also compare DrQ to a conventional DQN, whose reward function takes the form:
\begin{equation}
   r_{eng} = - (SOC_{t+1} - SOC_{target})^2 - \mathds{1}\left[ g_i(s_t,a_t) > 0 \right] 
\end{equation}
where we apply a constant step penalty for any constraint violation.

For each relevant constraint, we define a feed forward neural network to approximate $D_i$ with 2 hidden layers each with 10 neurons and sigmoid activation function.  We approximate the objective Q-function using a network of similar architecture. Table 2 includes the hyperparameters for the overall problem. The SPMeT model we use as a simulator is parameterized for a prismatic lithium nickel manganese cobalt oxide cell, different from the lithium iron phosphate cell used in our ECM case study. The episode simulation horizon is 1400 seconds.  For an optimal baseline, we use methods outlined in (\cite{Perez00}). Figure 8 shows the baseline control result.
\begin{figure}[]\label{fig:res1}
      \centering   
      \includegraphics[trim = 35mm 15mm 30mm 10mm, clip, width=0.95\textwidth]{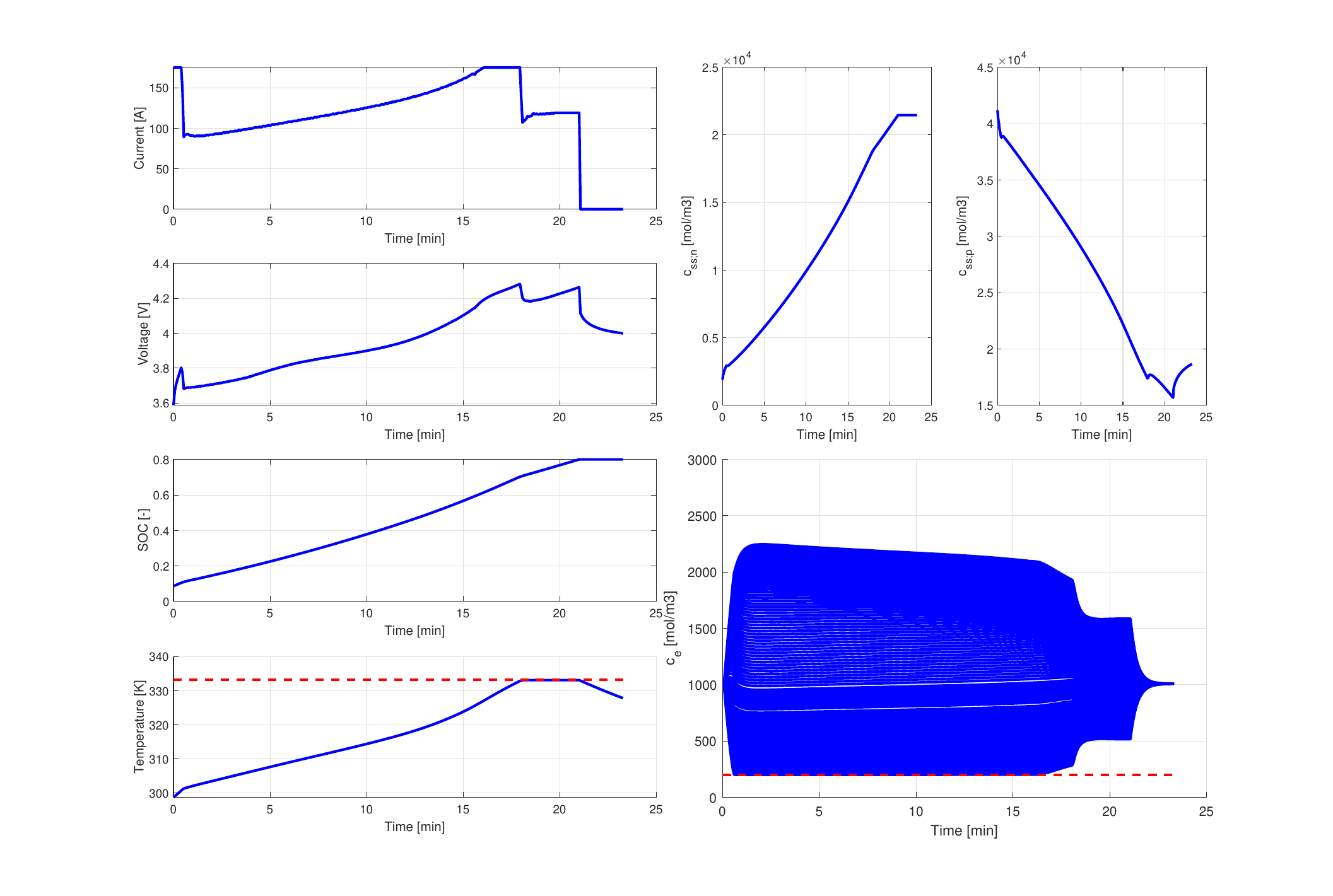}
      \caption{Baseline optimal control result.} 
      \label{figurelabel}
\end{figure}
Several meaningful insights can be taken from these baseline results.  First, the anode electrolyte constraint has the potential to be violated at almost every timestep.  Violation of this constraint can cause rapid aging or catastophic failure.  Second, the multitude of constraints provides a stronger challenge to our algorithm.   

\begin{table}[h!]
\caption{Relevant Parameters}
\label{sample-table}
\vskip 0.15in
\begin{center}
\begin{small}
\begin{sc}
\begin{tabular}{lcccr}
\toprule
Parameter & Description & Value & Units \\
\midrule
$\Delta t$    & Timestep & 4 &  $[s]$      \\
$\gamma$    & Discount Factor & 0.75 &  $[-]$      \\
$\epsilon$    & Exploration Prob. & 0.2 &  $[-]$      \\
$D_{\Xi;i}$    & Support Rad. & 1 &  $[]$      \\
$\beta$    & DRO Confidence & 0.9 &  $[-]$      \\
$\eta$    & CC Confidence & 0.05 &  $[-]$      \\
\bottomrule
\end{tabular}
\end{sc}
\end{small}
\end{center}
\vskip -0.1in
\end{table}

\subsection*{DrQ Results}
We simulate 10 independent runs of both DrQ and DQN, each with 25 episodes.  Figure 9 shows a comparison of the performance between DrQ and DQN, averaged across the runs. We evaluate the greedy policy performance at episodes 2, 5, 10, 15, 20, and 25 for computational purposes, given the SPMeT model is numerically expensive to simulate.  Relative to DrQ, the variance in the DQN performance is greater. Furthermore, the performance of DrQ is on average 31\% greater compared to DQN. One important note is that, to get the DQN baseline properly running, we had to add a fail-safe which set the input current to zero if the anode electrolyte concentration $c_e^-$ became too low.  DQN tended to max out the input current in the first several episodes of each run, which would rapidly deplete the electrolyte.  With a real battery cell, this would cause unsafe charging conditions and potential for catastrophic failure.  In simulation, however, this would simply terminate the code with numerical errors.

The safety advantages of DrQ can be seen in plots of constraint satisfaction.  Figures 10 and 11 show these results for the two active constraints, namely $c_e^-$ and $T$.  Figure 10 is particularly informative, since the electrolyte constraint is most often the dominant constraint.  Here, DrQ strongly attenuates the magnitude and frequency of constraint violation.  The temperature constraint is violated less for several reasons.  First, the temperature constraint only becomes active after a history of high input current. DrQ shows faster convergence to an optimal policy, which aggressively charges the battery.  Therefore, this constraint is violated more compared to DQN, which yields lower performing policies on average.  Nevertheless, the risk metric $\eta$ of the DrQ algorithm ($\eta = 5\%$) is validated within these experiments for all of the constraints, given that over all of the episodes and runs only 4.82\% of timesteps exhibited any constraint violation.  

Based on our data, the average DrQ episode took 361.58 seconds while the average DQN episode took 90.12 seconds.  Both simulate faster than real-time, which suggests that DrQ (and DQN) could run within on-board microcontrollers, even for complex dynamical systems.

\begin{figure}[b!]\label{fig:res1}
      \centering   
      \includegraphics[scale=0.46]{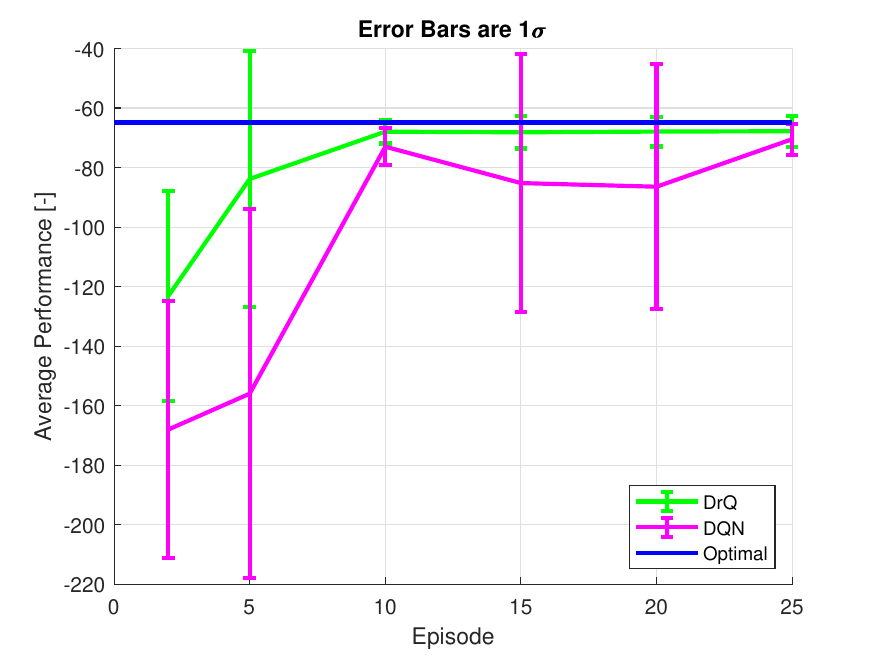}
      \caption{Greedy policy performance statistics over 10 runs of DrQ and DQN, based on the reward function defined by (\ref{eqn:rew}). } 
      \label{figurelabel}
\end{figure}
\begin{figure*}[ht!]\label{fig:res1}
      \centering   
      \includegraphics[trim = 10mm 1mm 5mm 5mm, clip, width=\textwidth]{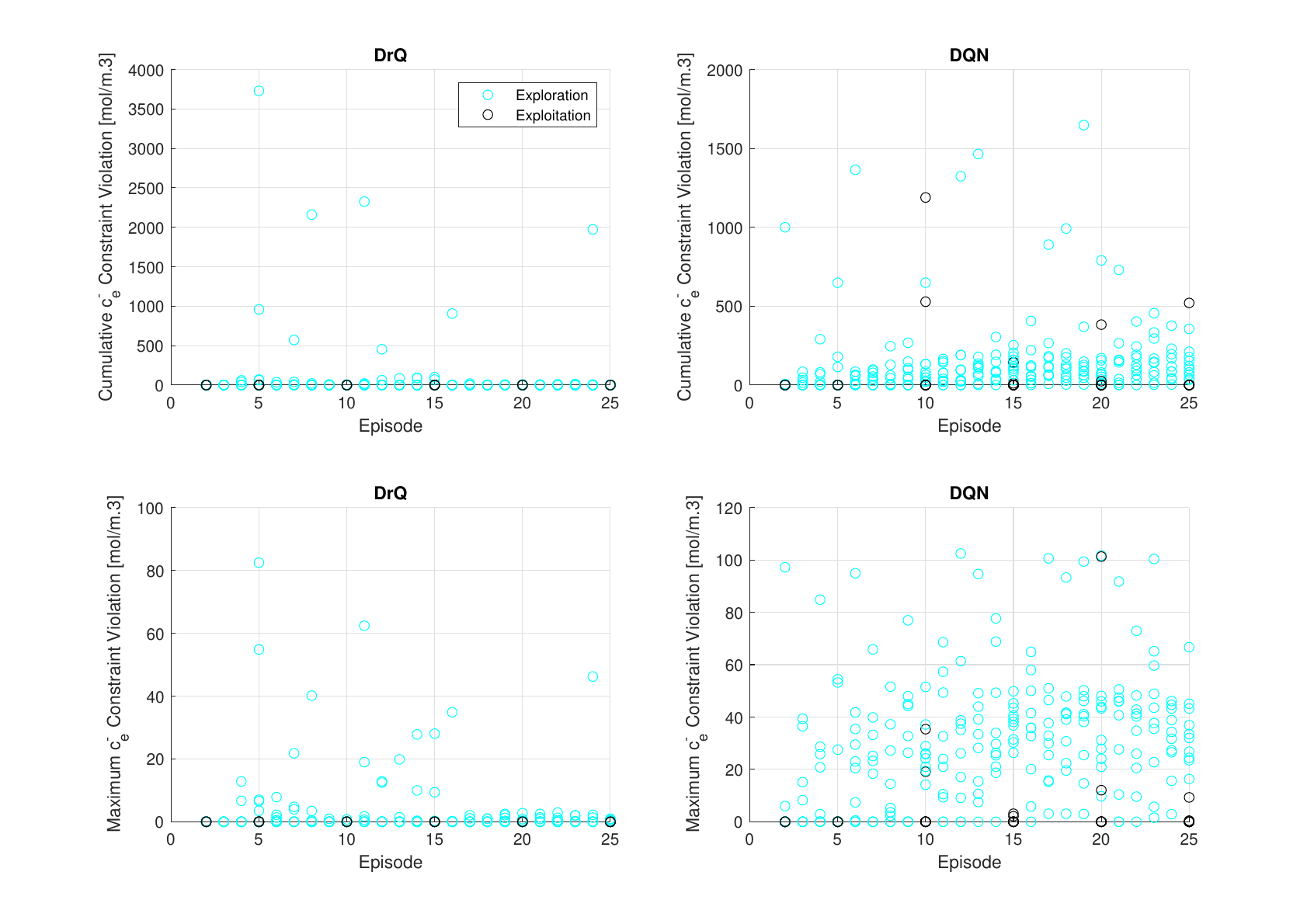}
      \caption{Safety statistics for the active $c_e^-$ constraint over 10 runs of DrQ and DQN, starting from the second episode.  The black ``exploration'' points correspond to the data obtained from the $\epsilon$-greedy policy. The cyan ``greedy'' points correspond to the greedy policy evaluated incrementally across each run.   } 
      \label{figurelabel}
\end{figure*}
\begin{figure*}[ht!]\label{fig:res1}
      \centering   
      \includegraphics[trim = 10mm 1mm 5mm 5mm, clip, width=\textwidth]{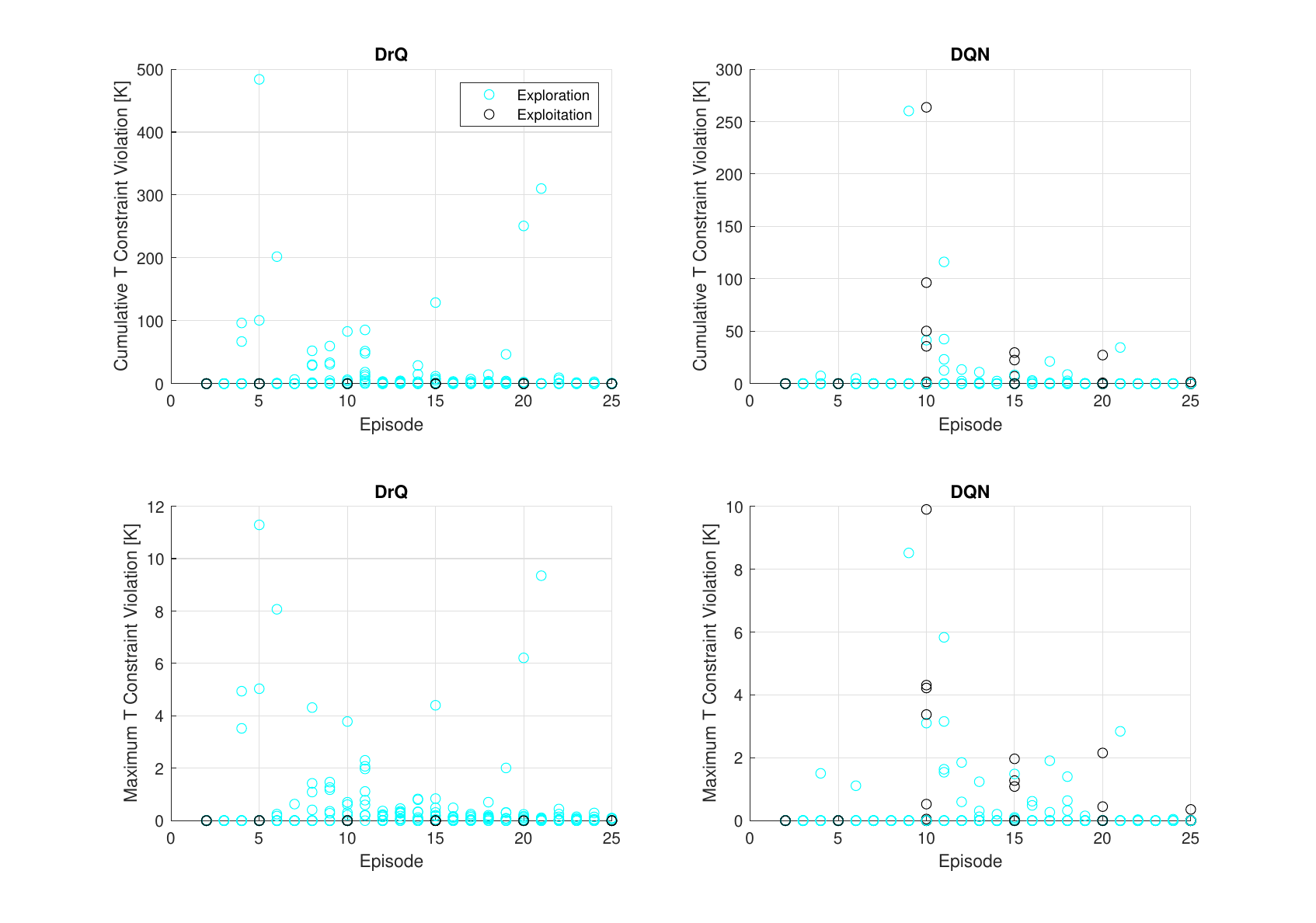}
      \caption{Safety statistics for the active $T$ constraint over 10 runs of DrQ and DQN, starting from the second episode.} 
      \label{figurelabel}
\end{figure*}

\bibliography{main}

\end{document}